\documentclass[
]{ceurart}

\usepackage{listings}
\usepackage{longtable}
\begin{document}

\copyrightyear{2026}
\copyrightclause{Copyright for this paper by its authors.
  Use permitted under Creative Commons License Attribution 4.0
  International (CC BY 4.0).}

\conference{CLEF 2026 Working Notes, 21 -- 24 September 2026, Jena, Germany}

\title{Fin-Analyst at FinMMEval 2026 Task 3: A Live Hybrid
Trading Agent with LLM Specialists and Rule-Based
Signals
}

\title[mode=sub]{<FinMMEval> at CLEF 2026}

\tnotemark[1]


\author[1]{Mohotarema Rashid}[%
orcid=0009-0003-2683-7191,
  email=MohotaremaRashid@my.unt.edu,
]
\cormark[1]

\author[2]{Lingzi Hong}[%
orcid=0000-0001-8412-8180,
  email=Lingzi.Hong@unt.edu,
]

\author[2]{Junhua Ding}[%
orcid=0000-0002-2129-1586,
  email=Junhua.Ding@unt.edu,
]

\author[2]{K.S.M.Tozammel Hossain}[%
orcid=0000-0003-0136-2145,
  email=Tozammel.hossain@unt.edu,
]

\address[1]{Department of Information Science, University of North Texas,
            Denton, TX, United States}
\address[2]{Department of Data Science, University of North Texas,
            Denton, TX, United States}

\cortext[1]{Corresponding author.}
\begin{abstract}
  Large language model (LLM) trading agents show promising performance in equity
markets, yet remain narrowly focused on US equities with little
evidence from live deployment. We present
Fin-Analyst, a hybrid agent for FinMMEval 2026 Task~3: an
eight-specialist LLM pipeline over news, SEC filings, fundamentals,
analyst forecasts, technical indicators, and social sentiment,
aggregated by a Meta-Agent for Tesla (TSLA), and a lightweight
rule-based three-signal vote for Bitcoin (BTC). On the final official
leaderboard (accessed 2026-07-05), Fin-Analyst ranks first of all
agents on TSLA with a $+13.51\%$ return, $+28.33$ points over
Buy-and-Hold (Sharpe 4.10, $88\%$ win rate), while the BTC vote ends
flat yet well above a sharply falling baseline. Relative to the interim
performance, the asset ranking reversed, indicating that short live
windows yield volatility-sensitive rankings. Ablation identifies
event-driven 8-K disclosures as the most influential TSLA signal.
Error analysis shows that the memoryless agents repeat wrong calls for
days at a time, and that the fixed-threshold BTC rules lost money by
trading on noise in a sideways market while the LLM pipeline gained
under similar conditions, motivating a memory-aware, LLM-based
successor for both assets.
\end{abstract}


\maketitle

\section{Introduction}
Financial markets function as vast information processing systems in which
security prices aggregate signals from heterogeneous sources~\cite{Fama1970,
Goldstein2021}. Therefore, trading in financial markets is inherently
data-driven. This data-driven approach relies on both structured numerical data, such as price, volume, accounting fundamentals
and unstructured textual data including news reports, regulatory filings,
analyst forecasts and social media discussions~\cite{Tetlock2007}. The rapid advancement of Artificial Intelligence (AI) and Machine Learning (ML) has further accelerated the adoption of algorithmic 
 and high-frequency trading systems that significantly automated trading operations and reduced direct human involvement~\cite{Hendershott2011}.
Yet classical algorithmic approaches like statistical arbitrage and
neural networks for time-series prediction have struggled to integrate
unstructured real-time data despite their proven predictive power~\cite{Cong2025,
Fischer2018}. The emergence of Large Language Models (LLMs) offers a promising
solution to these shortcomings, due to their ability to natively process both structured
and unstructured textual information and perform reasoning over heterogeneous sources of evidence~\cite{Wei2023}.
As a result, scholars and practitioners have begun adapting LLMs to
trading tasks~\cite{LopezLira2023}. A recent wave of LLM-based trading
agents includes news-driven~\cite{LopezLira2023}, reflection-driven~\cite{Yu2024},
debate-driven~\cite{Li2023}, and reinforcement-learning-driven~\cite{Koa2024}
systems that showed notable performance in backtested settings.

Despite this progress, two empirical limitations of existing LLM trading
agents exist. First, work has concentrated on US equities while
cryptocurrency markets remain substantially under-represented, and the
integration of such agents into live trading infrastructure has received
little attention. The FinMMEval Task~3 setting at CLEF~2026~\cite{FinMMEval2026,FinMMEvalTask3Overview2026}
brings these gaps into sharp focus: it replaces offline backtests with
a live, sequential, online evaluation protocol over both equity
(Tesla Inc.\ [TSLA]) and cryptocurrency (Bitcoin [BTC]) assets. This
setting directly motivates the multi-specialist agent developed in
this paper. In the above context, our contribution follows:

\begin{itemize}
\item A multi-agent LLM-based system for TSLA that combines news, SEC
      filings (8-K, 10-Q, 10-K), company fundamentals, analyst data,
      technical indicators, and social-media sentiment through a
      Meta-Agent aggregator.
\item A rule-based BTC strategy that uses price trend, the Crypto
      Fear \& Greed index, and momentum.
\end{itemize}

\begin{figure}[t]
  \centering
  \fbox{\includegraphics[width=0.96\linewidth]{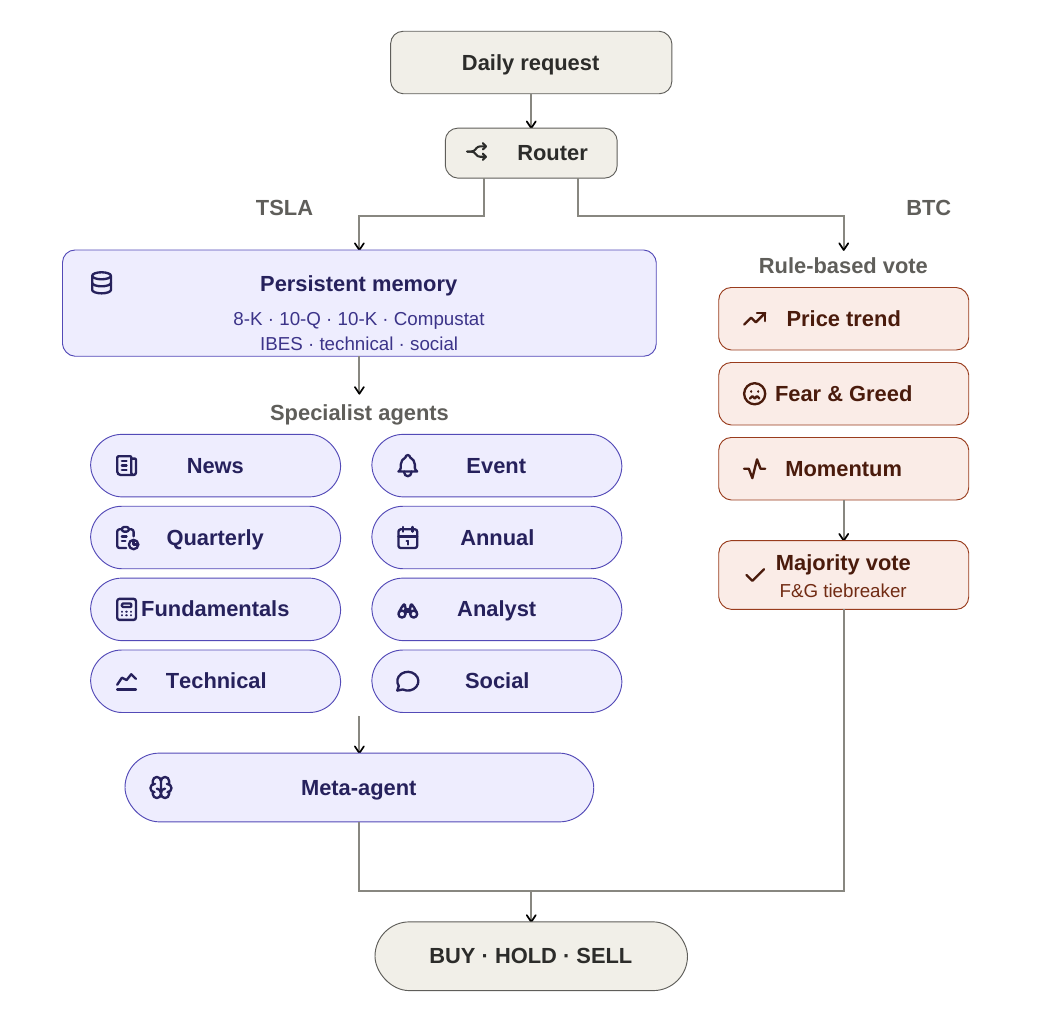}}
  \caption{Fin-Analyst architecture.}
  \label{fig:architecture}
\end{figure}

\section{Related Work}

LLMs have demonstrated notable capabilities in complex decision-making
tasks across multiple domains including healthcare, software engineering,
and scientific discovery~\cite{Park2023, Yu2024}. These highlight the
value of LLMs to address complex and sequential decision making
task~\cite{Schick2023}.

An emerging area of applying such frameworks to financial trading has
grown rapidly in recent years~\cite{Li2023, LopezLira2023, Takayanagi2025,
Yu2024, Zhang2024}. Scholars have shown that LLMs like GPT can
extract meaningful signals from news headlines~\cite{LopezLira2023}.
Zhang et al.~\cite{Zhang2024} developed a specialized agent that processes
multimodal data across news and technical indicators. These works
collectively highlight the effectiveness of LLM-based trading methods.

Beyond these simplified approaches, research also focuses on the memory
layers of LLMs to make effective trading decisions. Yu et al.~\cite{Yu2024}
organize sequential trading decisions by incorporating three layers of
memory: observation, summary, and reflection. Moreover, scholars use
summarization modules to distill long financial documents into concise summaries suitable for the LLM context window~\cite{Fatouros2024}.

In addition to pure in-context learning, several works train LLM agents
using outcome-driven feedback. For example, Koa et al.~\cite{Koa2024}
developed a self-reflective framework that applies proximal policy
optimization (PPO) to refine the agent's predictions based on historical
correct and incorrect outcomes.

Despite these advances, gaps remain in the existing LLM-based trading
literature. Evaluation of trading systems is predominantly confined to
US equity markets with limited attention to crypto securities~\cite{Ding2024},
and existing systems are rarely deployed in real-time market
scenarios~\cite{Qian2026}. The system developed in this paper addresses
these gaps through an eight-specialist agent architecture that operates
simultaneously on equity (TSLA) and cryptocurrency (BTC) under a live
online protocol, supported by a curated in-memory data pipeline loaded
once at startup.

\section{Methodology}
\label{sec:method}

\subsection{Task Formulation}
\label{sec:task}

We address Task~3 of FinMMEval at CLEF~2026, a daily online trading
task. Each day $t$, the evaluator sends a request containing the date
$d_t$, the latest price $p_t$, the price history $h_t$, a news bundle
$n_t$, and a momentum label $m_t$ for an asset
$s \in \{\text{TSLA}, \text{BTC}\}$. Our system returns a discrete
action $a_t \in \{\text{BUY}, \text{HOLD}, \text{SELL}\}$ for each
request.

\subsection{System Overview}
\label{sec:overview}

We propose a hierarchical agent with three layers: (i)~a persistent
memory of seven pre-processed corpora loaded at startup, (ii)~eight
role-conditioned LLM specialists, and (iii)~a Meta-Agent that
aggregates specialist verdicts into the final trading action. For
TSLA, the full hierarchy is used, for BTC the system bypasses the
LLM pipeline and applies a rule-based vote. Figure~\ref{fig:architecture} shows the overall architecture of our proposed system.

\subsection{Agent Specialization and Aggregation for TSLA}
\label{sec:specialists}

Each of our eight specialized agents consumes one data source: news
(today's bundle), 8-K filings (price-sensitive information), 10-Q
reports (quarterly disclosure), 10-K reports (annual disclosure),
Compustat fundamentals, IBES analyst consensus, technical indicators,
and social-media posts. Each specialist is implemented as a single
\textsc{gpt-4o-mini} call  that produces
an action, a confidence score, and a brief reasoning for the trading
decision. Table~\ref{tab:specialists} summarizes the specialists and
their inspiration in prior work. The Meta-Agent applies a
recency-weighted hierarchy in which today's news carries the highest
weight and may override consensus when its confidence exceeds a
predefined threshold. Each disclosure-based specialist always operates
on the most recent filing dated on or before the decision date;
filings are carried forward, so on days with no new 8-K, 10-Q, or
10-K the corresponding specialist re-reads the latest prior filing
rather than abstaining. This ensures every specialist returns a
verdict every day.

\begin{table}[t]
\centering
\caption{The eight TSLA specialists, their data sources, and prior
inspiration.}
\label{tab:specialists}
\small
\begin{tabular}{llll}
\toprule
\textbf{Specialist} & \textbf{Data Source} & \textbf{Prior Inspiration}
& \textbf{Proxy for} \\
\midrule
News         & Organizer-provided bundle & \cite{LopezLira2023} & News sentiment \\
Event        & 8-K filings               & \cite{Rawson2023}    & Price-sensitive info. \\
Quarterly Financial Disclosure     & 10-Q               & \cite{Yu2024}        & Quarterly performance \\
Annual Financial Disclosure     & 10-K                      & \cite{Yu2024}        & Annual performance \\
Fundamentals & Compustat                 & \cite{Asness2013}    & Financial ratios \\
Analyst      & IBES consensus            & \cite{Zhang2024}     & Analyst forecasts \\
Technical    & MACD, RSI, BB, ADX        & \cite{Zhang2024}     & Market momentum \\
Social       & \textit{r/wallstreetbets} & \cite{Koa2024}       & Social media sentiment \\
\bottomrule
\end{tabular}
\end{table}

\subsection{Cryptocurrency Logic for BTC}
\label{sec:btc}

For BTC, we use a rule-based three-signal majority vote over the price
trend, the Crypto Fear \& Greed Index, and the organizer-provided
momentum label, with ties broken by the Fear \& Greed score. The
Crypto Fear \& Greed Index is a daily sentiment indicator for the
cryptocurrency market, published by alternative.me. It aggregates
several market signals including social-media activity, market
momentum, and Bitcoin dominance into a single score between 0 and
100, where lower values indicate ``Extreme Fear'' and higher values
indicate ``Extreme Greed.'' The index is widely used, with extreme
fear historically associated with long opportunities (buy) and extreme
greed with increased correction risk~\cite{Wang2024}. In our BTC
pipeline, the index is queried at decision time and mapped to a
directional vote: BUY when the score is $\geq 60$, SELL when
$\leq 40$, and HOLD otherwise. It also serves as the tiebreaker when
the price-trend and momentum signals disagree.

The price-trend signal is computed from the organizer-provided price
history $h_t$ as the percentage change between the first and last
price in the window,
$r_t = (p_{\text{last}} - p_{\text{first}})/p_{\text{first}} \times 100$.
It votes BUY when $r_t > +0.5\%$, SELL when $r_t < -0.5\%$, and HOLD
otherwise. The momentum label maps bullish$\rightarrow$BUY and
bearish$\rightarrow$SELL. The final action is the majority of the
three votes, with the Fear \& Greed score breaking ties (BUY if
$\geq 50$, else SELL).

\subsection{Deployment}
\label{sec:deployment}

Our system is deployed as a containerized FastAPI service on Hugging
Face Spaces, with all persistent corpora loaded into memory at
startup. On each request the router dispatches by asset. For TSLA,
the eight specialists are invoked sequentially as blocking
\textsc{gpt-4o-mini} calls via the OpenAI Python SDK, each constrained
to JSON-only output with a per-agent token cap, after which the
Meta-Agent is called once nine LLM calls per decision. To bound
latency and cost, verdicts are cached by an MD5 hash of the prompt, so
specialists reading unchanged low-frequency signal cards (e.g., the
same most-recent 8-K on consecutive days) reuse a cached verdict
rather than issuing a new API call. Any specialist or endpoint exception returns a HOLD verdict,
ensuring compliance with the task's no-retry protocol and addressing
the observation of Ding et al.~\cite{Ding2024} that integration of
LLM trading agents with live trading systems is rarely demonstrated
in prior work.

\subsection{Implementation Details}
\label{sec:impl}

We use the \textsc{gpt-4o-mini} model for all specialists and the
Meta-Agent with temperature~0.1. See
Table~\ref{tab:prompts}. Prompts follow the role-play conditioning style of
Yu et al.~\cite{Yu2024}. The full prompts are listed in \ref{app:prompts}.

\section{Dataset}
\label{sec:dataset}

We evaluate our proposed system on the CLEF~2026 Task~3 dataset
(\texttt{TheFinAI/CLEF\_Task3\_Trading}), which spans two assets: Tesla (TSLA, equity)
and Bitcoin (BTC, cryptocurrency). Each daily record contains the close
price, a pre-summarized news bundle, a momentum label, and the realized
next-day price change. The action space is \{BUY, HOLD, SELL\}.
Table~\ref{tab:dataset} summarizes the key dataset properties.

\begin{table}[t]
\centering
\caption{CLEF~2026 Task~3 evaluation windows.}
\label{tab:dataset}
\small
\begin{tabular}{ll}
\toprule
\textbf{Property} & \textbf{Value} \\
\midrule
Source        & \texttt{TheFinAI/CLEF\_Task3\_Trading} \\
Asset classes & Equity (TSLA), Cryptocurrency (BTC) \\
Fields        & date, price, news, momentum, future price change \\
Action space  & \{BUY, HOLD, SELL\} \\
\midrule
\multicolumn{2}{l}{\textit{Offline backtest (historical dataset)}} \\
Period        & 2025-08-01 to 2026-05-10 \\
TSLA          & 194 trading days (\$302.63--\$489.88) \\
BTC           & 283 trading days (\$62{,}754--\$124{,}798) \\
\midrule
\multicolumn{2}{l}{\textit{Live challenge (Agent Market Arena)}} \\
Period        & from 2026-05-11 (leaderboard accessed 2026-07-05) \\
TSLA          & 33 trading days (through 2026-06-26) \\
BTC           & 50 days (through 2026-06-29) \\
\bottomrule
\end{tabular}
\end{table}
\section{Results}
\label{sec:results}

\subsection{Evaluation metrics}
\label{sec:metrics}
We adopt four portfolio metrics standard in the LLM trading agent
literature~\cite{Koa2024,Yu2024,qian2025whenagentstrade}.

\textbf{Cumulative Return (CR)} measures total portfolio growth over
an evaluation window of length $T$:
\begin{equation}
  \mathrm{CR} = \frac{P_T - P_0}{P_0},
\end{equation}
where $P_0$ and $P_T$ are portfolio values at the start and end of the
window.

\textbf{Alpha versus Buy-and-Hold} ($\alpha_{\text{BH}}$) captures the
return the agent generates beyond a passive benchmark on the same
asset:
\begin{equation}
  \alpha_{\text{BH}} = \mathrm{CR}_{\text{agent}}
                      - \mathrm{CR}_{\text{BH}}.
\end{equation}

\textbf{Sharpe Ratio (SR)} quantifies risk-adjusted return as the
ratio of mean excess return to its standard deviation:
\begin{equation}
  \mathrm{SR}
    = \frac{\mathbb{E}[R_p] - R_f}{\sigma_p},
\end{equation}
where $R_p$ is the realized portfolio return, $R_f$ the risk-free rate
(set to zero following Qian et al.~\cite{qian2025whenagentstrade}),
and $\sigma_p$ the standard deviation of excess returns.

\textbf{Win Rate (WR)} measures the proportion of executed trades that
closed profitably:
\begin{equation}
  \mathrm{WR} = \frac{N_{\text{wins}}}{N_{\text{trades}}},
\end{equation}
complementing the Sharpe ratio by isolating directional accuracy from
magnitude effects~\cite{Bailey2014}.
\subsection{Live evaluation}
\label{sec:live}
\begin{figure}[t]
  \centering
  \includegraphics[width=\linewidth]{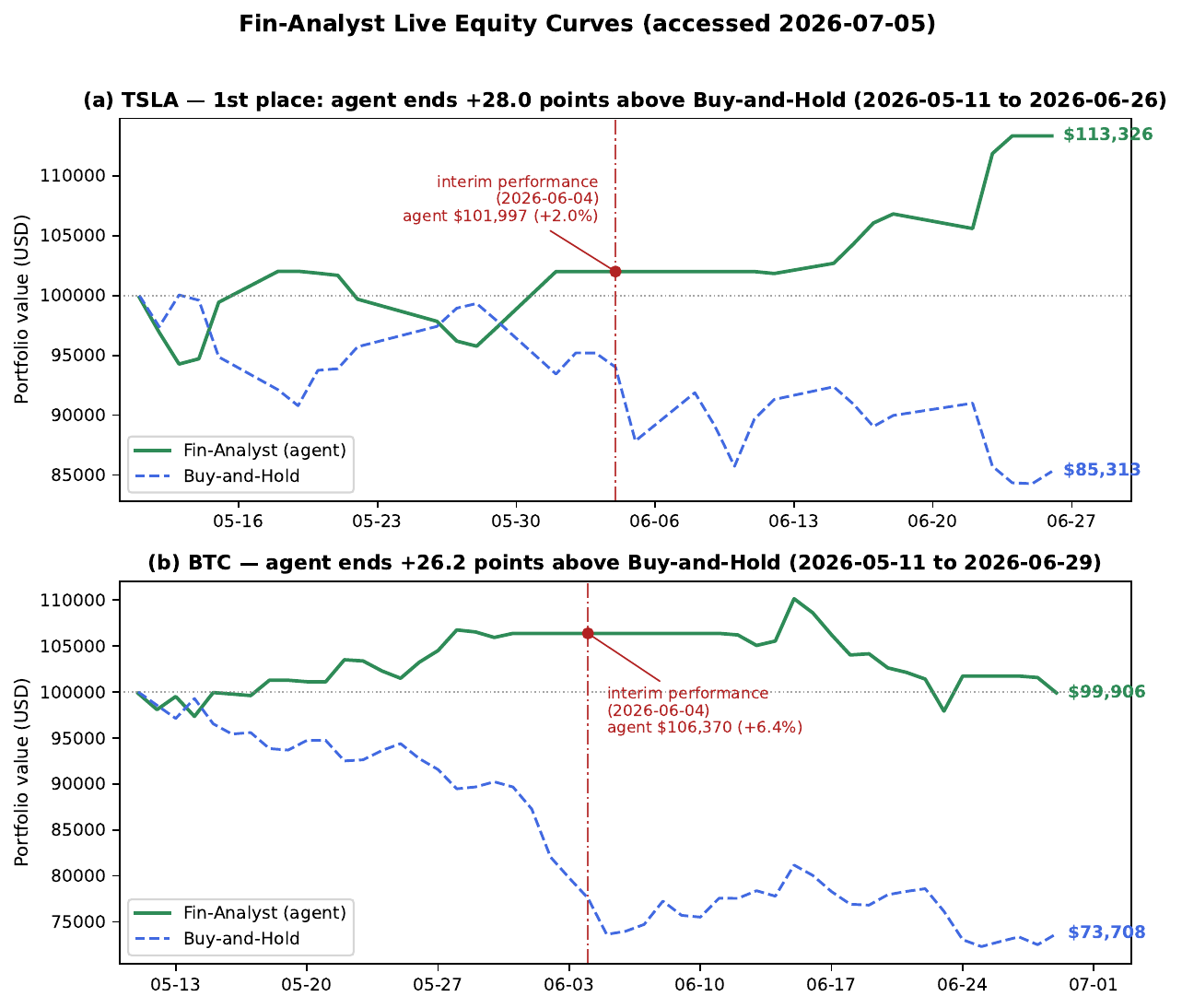}
  \caption{Live equity curves over the  official window (TSLA
  2026-05-11 to 2026-06-26; BTC 2026-05-11 to 2026-06-29),
  reconstructed from the arena's official per-day trading log. The
  dash-dotted vertical line marks the interim working-notes snapshot
  date (2026-06-04), at which the corrected series stood at $+2.0\%$
  (TSLA) and $+6.4\%$ (BTC).}
  \label{fig:equity}
\end{figure}

We evaluate Fin-Analyst on the live Agent Market
Arena~\cite{qian2025whenagentstrade} over the FinMMEval Task~3
challenge~\cite{FinMMEvalTask3Overview2026}, which started 2026-05-11.
The original working-notes discussion was based on an interim live
leaderboard performance (2026-06-04); this version reports the update accessed 2026-07-05, with
TSLA data available through 2026-06-26 and BTC through 2026-06-29.
Returns are net of transaction fees against a passive
Buy-and-Hold baseline; Table~\ref{tab:results} and
Figure~\ref{fig:equity} report the outcome.

\begin{table}[t]
\centering
\caption{Live evaluation results for Fin-Analyst on the
FinMMEval Task~3 Agent Market Arena (accessed 2026-07-05). Returns
are net of fees; ``Vs B\&H'' is return relative to the passive
Buy-and-Hold baseline. TSLA metrics are from the official leaderboard
(data through 2026-06-26); BTC return and alpha are computed from the
official per-day trading log through 2026-06-29 via Eqs.~(1)--(2),
with rank as displayed on the
leaderboard at access}
\label{tab:results}
\small
\renewcommand{\arraystretch}{1.2}
\begin{tabular}{@{}lrrrrc@{}}
\toprule
\textbf{Asset} & \textbf{Return} & \textbf{Vs B\&H} & \textbf{Sharpe} & \textbf{Win Rate} & \textbf{Rank} \\
\midrule
TSLA & $+13.51\%$ & $+28.33\%$ & $4.10$ & $88\%$ & 1st / gold \\
BTC  & $-5.30\%$  & $+17.63\%$ & $-1.09$ &  $36\%$ & 13th \\
\bottomrule
\end{tabular}
\end{table}

On TSLA, the multi-specialist LLM pipeline returns $+13.51\%$ against
a $-14.7\%$ Buy-and-Hold ($+28.33$ points, Sharpe $4.10$, $88\%$ win
rate), finishing first of all submitted agents. On BTC, the
rule-based pipeline ends essentially flat at $-5.30\%$, yet $+17.63$
points above a Buy-and-Hold that lost roughly a quarter of its value:
its early short positioning captured the May drawdown before the lead
decayed in late June (ranked 13th at access). Figure~\ref{fig:equity}
makes the window sensitivity explicit: at the interim snapshot date
(2026-06-04) the corrected series show TSLA at only $+2.0\%$ and BTC
at $+6.4\%$, so the TSLA gold-medal run occurred almost entirely
after the working-notes cutoff and the interim asset ranking reversed
by the final window. This reversal is our central live result: over
short, volatile windows, live agent rankings are unstable, and
conclusions drawn from any single snapshot should be treated with
caution.
\subsection{Offline backtest and ablation}
\label{sec:offline}
To test our proposed system beyond the short live window, we backtest
over the historical CLEF Task~3 dataset (2025-08-01 to 2026-05-10) and isolate each specialist's contribution by
ablation.  Following common practice~\cite{Qian2026}, we
report cumulative return (CR), Sharpe (SR), annualized return (AR),
maximum drawdown (MDD), win rate (WR), and alpha to Buy-and-Hold
($\alpha_{\text{BH}}$). 

\begin{table}[t]
\centering
\caption{Offline backtest on TSLA, NewsOnly is
the Lopez-Lira and Tang~\cite{LopezLira2023} single-prompt news
baseline.}
\label{tab:tsla}
\small
\begin{tabular}{lrrrrrrr}
\toprule
\textbf{Strategy} & \textbf{CR} & $\alpha_{\text{BH}}$ & \textbf{SR}
& \textbf{AR} & \textbf{MDD} & \textbf{WR} & $N_{\text{tr}}$ \\
\midrule
Buy \& Hold              & $+0.379$ & $0.000$  & $+0.96$ & $+0.318$ & $0.299$ & $0.37$ & 293 \\
Always HOLD              & $0.000$  & $-0.379$ & $0.00$  & $0.000$  & $0.000$ & $0.00$ & 0   \\
Random ($\sigma{=}42$)   & $+0.417$ & $+0.039$ & $+1.13$ & $+0.350$ & $0.290$ & $0.28$ & 211 \\
Momentum                 & $-0.434$ & $-0.812$ & $-1.22$ & $-0.387$ & $0.652$ & $0.32$ & 289 \\
NewsOnly                 & $+0.217$ & $-0.162$ & $\mathbf{+0.94}$ & $+0.184$ & $0.143$ & $0.13$ & 77 \\
\textbf{Fin-Analyst}     & $\mathbf{+0.256}$ & $-0.122$ & $+0.73$ & $\mathbf{+0.217}$ & $0.254$ & $0.32$ & 293 \\
\bottomrule
\end{tabular}
\end{table}

\begin{table}[t]
\centering
\caption{Offline backtest on BTC}
\label{tab:btc}
\small
\begin{tabular}{lrrrrrrr}
\toprule
\textbf{Strategy} & \textbf{CR} & $\alpha_{\text{BH}}$ & \textbf{SR}
& \textbf{AR} & \textbf{MDD} & \textbf{WR} & $N_{\text{tr}}$ \\
\midrule
Buy \& Hold              & $-0.315$ & $0.000$  & $-0.68$ & $-0.277$ & $0.497$ & $0.49$ & 294 \\
Always HOLD              & $0.000$  & $+0.315$ & $0.00$  & $0.000$  & $0.000$ & $0.00$ & 0   \\
Random                   & $+0.818$ & $+1.133$ & $+1.85$ & $+0.669$ & $0.148$ & $0.36$ & 211 \\
Momentum                 & $-0.062$ & $+0.254$ & $+0.04$ & $-0.053$ & $0.486$ & $0.50$ & 294 \\
\textbf{Fin-Analyst (rule-based)} & $\mathbf{+0.228}$ & $\mathbf{+0.543}$ & $\mathbf{+0.66}$ & $\mathbf{+0.193}$ & $0.255$ & $0.52$ & 294 \\
\bottomrule
\end{tabular}
\end{table}

In the backtest, the full TSLA system produces $+25.63\%$ CR (Sharpe
$+0.73$): the highest annualized return among LLM-based strategies and
$+3.9$ points above the  NewsOnly
baseline~\cite{LopezLira2023}, indicating the multi-specialist
architecture, rather than news alone, drives the result. NewsOnly
attains a marginally higher Sharpe ($+0.94$) by abstaining on most
days, a different trade-off between trading frequency and risk-adjusted
return~\cite{Bailey2014}. On BTC, the rule-based vote returns
$+22.82\%$ against a $-31.53\%$ Buy-and-Hold
($\alpha_{\text{BH}}=+54.34$ points) exceeds the best per-agent
BTC alpha in prior benchmarks~\cite{Qian2026}. The gap between these
backtest figures and the live result, most pronounced on BTC,
is itself consistent with the backtest-to-live degradation reported by
Qian et al.~\cite{qian2025whenagentstrade}.

\begin{table}[t]
\centering
\caption{ Ablation on TSLA.
$\Delta$-CR is the change in cumulative return when the named
specialist is disabled; a larger negative $\Delta$ indicates a greater
positive contribution.}
\label{tab:ablation}
\small
\begin{tabular}{lrrrrl}
\toprule
\textbf{Disabled} & \textbf{CR} & $\boldsymbol{\Delta}$\textbf{-CR}
& \textbf{SR} & $\boldsymbol{\Delta}$\textbf{-SR} & \textbf{Contribution} \\
\midrule
Event (8-K)                 & $-0.050$ & $\mathbf{-30.65}$pp & $+0.05$ & $-0.68$ & critical \\
Quarterly (10-Q)            & $+0.113$ & $-14.35$pp          & $+0.44$ & $-0.30$ & critical \\
News (daily bundle)         & $+0.139$ & $-11.74$pp          & $+0.49$ & $-0.24$ & critical \\
Annual (10-K)               & $+0.281$ & $+2.42$pp           & $+0.78$ & $+0.05$ & marginally negative \\
Fundamentals (Compustat)    & $+0.286$ & $+2.95$pp           & $+0.79$ & $+0.06$ & marginally negative \\
\bottomrule
\end{tabular}
\end{table}

The ablation (Table~\ref{tab:ablation}) yields several findings. The
Event (8-K) specialist is the  most influential signal consistent with the price impact of material
corporate disclosures~\cite{Rawson2023,Fama1970}. The Earnings (10-Q)
and News specialists are equal secondary drivers ($-14.35$ and
$-11.74$ points). The low-frequency 10-K and Compustat Fundamentals
specialists marginally hurt performance, which suggests that annual signals add
noise rather than signal at the daily-decision
horizon~\cite{Ramiah2015}.

\section{Discussions, Limitations and Error Analysis}
\label{sec:discussion}

Our proposed system outperforms the Buy-and-Hold baseline on both
assets in the final official window. The outcome, however, reversed
between the interim snapshot and the final window: the multi-specialist
LLM pipeline, flat at 2026-06-04, captured the late-June TSLA rally
and finished first, while the rule-based BTC vote, which led the
interim board, decayed to flat. Such rank instability over short,
volatile live windows is consistent with evidence that LLM trading
advantages are regime-sensitive and deteriorate under longer and
broader evaluation~\cite{Li2025finsaber}, with the adaptive,
regime-dependent nature of market efficiency~\cite{Lo2004}, and with
the backtest-to-live degradation reported by Qian et
al.~\cite{qian2025whenagentstrade}. For BTC, our rule based
three-signal majority vote was defensive rather than
return-generating: it preserved capital ($+26.2$ points of alpha) in
a market that fell by roughly $26\%$, consistent with evidence that
simple momentum and technical signals carry real predictive power in
cryptocurrency markets~\cite{Liu2021} and that the Fear \& Greed
Index aligns with short-horizon BTC returns~\cite{Wang2024}.

Moreover, the following limitations qualify these results and mark
where the next gains are likely to come from:
\begin{itemize}

\item \textbf{Short, volatile live window.} Both assets were
evaluated over fewer than two months of live trading in a
high-volatility regime; the interim-to-final rank reversal shows that
such windows cannot separate skill from regime
luck~\cite{Li2025finsaber,White2000}, so the live rankings reported
here should be read as indicative rather than conclusive.

\item \textbf{Multi-source orchestration requires careful aggregation.}
In the offline backtest, eight specialists pulling from heterogeneous sources appeared to introduce
noise in daily trading decisions, and our recency-weighted Meta-agent does not
fully discount the slower channels, which echoes Ding et al.'s
observation that naive scaling does not improve trading performance
without careful aggregation~\cite{Ding2024}. The strong final TSLA
window shows the ensemble can capture a favorable regime, but the
aggregation itself remains hand-tuned.

\item \textbf{The daily news signal dominates everything else.} Today's
news was clearly our highest-information reflection, consistent with
prior news-driven LLM trading task~\cite{LopezLira2023}. Incorporating slow
moving inputs such as the 10-K disclosure and social media sentiment through
the same vote appears to have diluted that opportunity rather than
reinforced it.

\item \textbf{An LLM-based BTC agent is the natural next step.} The
rule-based vote led the interim leaderboard but decayed to flat as
the regime turned, because its fixed thresholds cannot adapt. Given
that the multi-specialist LLM pipeline ultimately won the TSLA
track, extending it to BTC with crypto-native inputs (on-chain
activity, funding rates, crypto news) could pair the vote's
defensiveness with adaptive positioning.

\item \textbf{Memory engineering is essential.} Prior work shows that
layered  memory~\cite{Yu2024,Zhang2024} and reflection-driven
self-correction~\cite{Shinn2023} significantly improve agent
performance. We use a stateless prompt per request and store nothing
across time, which might be an obvious gap to close in future iterations.

\item \textbf{Reasoning engineering might improve the result.} Each specialist is a
single forward pass with no prompt engineering
~\cite{Wei2022}, no debate-style
aggregation~\cite{Du2023}, and no fine-tuning of the underlying
model. We conclude that these are standard paths in stronger financial
agents and would be our next focus.

\item \textbf{Our model may simply be too small.} We used
\textsc{gpt-4o-mini} for cost and latency . A stronger
foundation model such as GPT-4o, or a domain-fine-tuned variant,
could significantly shift performance before any of the other changes
above are made.

\item \textbf{We did not perform statistical significance testing.}
The performance gap between our system's returns and the baseline may not be
distinguishable from random variation~\cite{White2000}. Any
performance claim, positive or negative, needs proper hypothesis
testing to stand.

\item \textbf{Limited SOTA comparison.}
We benchmark against the single-prompt news baseline of Lopez-Lira
and Tang~\cite{LopezLira2023}, which isolates the contribution of
the multi-specialist architecture over a news-only LLM. However, we
do not yet compare against published multi-agent systems such as
FinMem~\cite{Yu2024} or FinAgent~\cite{Zhang2024}, which would
position our pipeline relative to the broader field. Extending the
evaluation to these systems is a priority for future work.

\end{itemize}

As a result, we do not claim that these live rankings are
definitive. What we offer is a
documented effort of what a multi-source LLM trading agent looks
like without memory, advanced reasoning, a larger model, or rigorous
evaluation, together with a clear list of where the next gains are
likely to come from.

\subsection{Error analysis}
\label{sec:error_analysis}
We study when the system made money and when it lost money, using the
arena's official day-by-day trading logs. The TSLA log covers 2026-05-11
to 2026-06-26 with 33 trading days, and the BTC log covers 2026-05-11 to
2026-06-29 with 50 days. A day counts as a hit when the position pointed
in the same direction that the price moved on the following day.
Table~\ref{tab:error_attribution} summarises the attribution.

\begin{table}[t]
\centering
\caption{Day-level error attribution computed from the official per-day
trading logs. Hit rates are position-vs-next-move measures and differ from
the arena's trade-level win rate.}
\label{tab:error_attribution}
\small
\renewcommand{\arraystretch}{1.2}
\begin{tabular}{@{}lcc@{}}
\toprule
 & \textbf{TSLA} & \textbf{BTC} \\
\midrule
Final return (net of fees)      & $+13.51\%$ & $-5.30\%$ \\
Acted days / exposure           & 19/33 (58\%) & 31/50 (62\%) \\
Hit rate (acted days)           & 0.632 & 0.387 \\
Long days: hit / total PnL      & 0.75 / $-1.55$\,pp & 0.17 / $-6.50$\,pp \\
Short days: hit / total PnL     & 0.60 / $+15.17$\,pp & 0.44 / $+7.71$\,pp \\
Max equity drawdown             & $-6.1\%$ & $-11.1\%$ \\
\bottomrule
\end{tabular}
\end{table}

Both markets fell during the evaluation window. TSLA lost $14.7\%$ and BTC
lost $26.3\%$ under Buy-and-Hold. Every point of profit came from short
positions, while long positions lost money on both assets. The system
performed well in two situations. In the first situation, the market was
clearly falling and negative news kept arriving, so the system stayed
short and profited. This happened on TSLA from 05-12 to 05-15 with a gain
of $+5.3\%$ and on BTC from 05-12 to 05-30 with a gain of $+8.3\%$, which
is consistent with the finding that LLM predictability concentrates in bad
news~\cite{LopezLira2023}. In the second situation, the TSLA pipeline
timed small ups and downs correctly in mid-June and gained $+9.6\%$ even
though the price barely moved. This stretch produced the winning margin,
and it happened after the interim snapshot.

We divide error analysis into four group by following error taxonomy of LLM based trading literature ~\cite{DAmico2026}. The first group is the bad first day.
On day one the system bought on both assets and lost, with TSLA falling
$2.92\%$ on its worst day and BTC falling $1.8\%$. The reason is that the
system starts with no history and no context, so its first decision is
effectively a blind guess.

The second group is holding on to a wrong bet. TSLA stayed short for six
days in a row while the price rose $6\%$ between 05-21 and 05-28, and this
stretch caused the maximum drawdown of $6.1\%$. BTC repeated the same
mistake by shorting a rising market on its final two days and losing
$1.8\%$. The reason is that the system has no memory, so it cannot
recognise that it has been wrong for several consecutive days and it keeps
repeating the same call. This failure is the trading analogue of the
action-inversion errors reported for financial text
classifiers~\cite{DAmico2026}.

The third group is trading on noise, and it produced the largest loss.
BTC lost $8.1\%$ between 06-11 and 06-22 while the market moved only
$+1.3\%$, and this single stretch wiped out the lead it had built earlier.
The reason is that the trading rules use a threshold of $\pm 0.5\%$, but
BTC normally moves 2 to 3 percent in a day. The rules therefore kept
flipping direction on random wiggles, and they issued buy signals only
after prices had already peaked, so only one of six buy decisions was
correct~\cite{Liu2021}.

The fourth group is sitting out big moves. BTC stayed flat while the
market fell $15.8\%$, because its three signals disagreed and the tie
defaulted to HOLD. TSLA stayed flat during a $4.0\%$ drop, because no news
arrived to trigger a trade. These mistakes are safe in the sense that no
money was lost, but the profit that was available in those periods was
missed.

The clearest lesson comes from comparing the two pipelines over the same
near-flat weeks. The LLM pipeline gained $+9.6\%$ on TSLA while the
fixed-threshold rules lost $8.1\%$ on BTC. Reading the news day by day
works in sideways markets, whereas fixed thresholds do not. This contrast
directly motivates the LLM-based BTC agent and the memory layer proposed
in Section~\ref{sec:future}.

\section{Future Work}
\label{sec:future}
Our proposed system relies on static prompting and a hand-tuned
aggregation rule for TSLA, and on a rule-based BTC vote with fixed
decision thresholds ($\pm 0.5\%$ price trend, Fear~\&~Greed
$40/60$), neither of which learns from realized trading outcomes.
Future directions include (i)~refine specialist prompts and
Meta-Agent weights via proximal policy optimization against next-day
returns, (ii)~replace single-pass aggregation with structured
multi-agent debate, (iii)~introduce a regime-conditioned memory
layer that discounts low-frequency annual signals during
high-volatility periods, extending Yu et al.~\cite{Yu2024}, and
(iv)~make the BTC vote regime-aware by re-fitting the price-trend
and Fear~\&~Greed thresholds on a rolling window and adding a
moving-average trend filter to exit short positions when the market
turns, addressing the late-June decay observed in
Section~\ref{sec:live}. Together, these directions outline a
learned, debate-driven, and memory-aware successor to the present
prompt-engineered system, and align with the open challenges of
stronger backbones, rigorous evaluation, and live deployment
identified for LLM trading agents at large~\cite{Ding2024}.

\section*{Declaration on Generative AI}
   During the preparation of this work the author(s) used AI tools such as Claude to assist in figure creation, and writing of some of the sections. The authors developed all core ideas, methods, analyses, and conclusions. The final content reflects the authors' independent scholarly contributions. After using this tool/service, the author(s) reviewed and edited the content as needed and take(s) full responsibility for the content of the published article. 

\bibliography{sample-ceur}

\appendix
\renewcommand{\thesection}{Appendix}

\section{Specialist Prompts}
\label{app:prompts}

Table~\ref{tab:prompts} lists the system and user prompts for the
eight specialist agents and the meta agent. All agents are called
through \textsc{gpt-4o-mini} at temperature 0.1 with JSON-only output
of the form \texttt{\{action, confidence, reasoning\}}. If any
specialist call errors out, the system falls back to a default HOLD
verdict.

\renewcommand{\arraystretch}{1.3}
\footnotesize
\setlength{\tabcolsep}{4pt}
\begin{longtable}{@{}p{0.14\linewidth} p{0.46\linewidth} p{0.34\linewidth}@{}}
\caption{System and user prompts for the eight TSLA specialists and
the meta agent. Placeholders in braces (e.g.\ \{date\},
\{news\_text\}) are filled in at inference time with data drawn from
the corresponding pre-processed signal card.}
\label{tab:prompts} \\
\toprule
\textbf{Agent} & \textbf{System prompt} & \textbf{User prompt template} \\
\midrule
\endfirsthead

\multicolumn{3}{l}{\emph{Table~\ref{tab:prompts} (continued)}} \\
\toprule
\textbf{Agent} & \textbf{System prompt} & \textbf{User prompt template} \\
\midrule
\endhead

\bottomrule
\endlastfoot

\textbf{News} &
You are the NEWS Specialist for daily TSLA trading. Strong BUY
(conf 0.80--0.90) on earnings beats, positive product launches,
analyst upgrades, insider buying. Strong SELL on misses, lawsuits
naming specific dollar amounts, recalls, executive departures,
regulatory probes, insider selling. Multiple negatives compound.
Never default to HOLD when a clear catalyst is present; reasoning
must cite the specific phrase. Output JSON. &
Date: \{date\} \newline
TSLA News for today: \{news\_text\} \newline
Judge market impact for next 1--3 days. \\
\midrule

\textbf{Event} &
You are an Event Specialist analyzing the most recent SEC 8-K filing.
Earnings + positive impact $\rightarrow$ BUY 0.85; earnings +
negative $\rightarrow$ SELL 0.85; M\&A $\rightarrow$ BUY 0.70;
CEO/CFO departure $\rightarrow$ SELL 0.65; routine management
changes $\rightarrow$ BUY 0.40; regulatory or legal trouble
$\rightarrow$ SELL 0.70. Output JSON. &
Filing date: \{date\} \newline
Event: \{event\_category\} \newline
Market impact: \{market\_impact\} \newline
Sentiment: \{sentiment\}, Tone: \{tone\} \newline
Summary: \{summary[:300]\} \\
\midrule

\textbf{Earnings} &
You are an Earnings Specialist using 10-Q commentary + IBES surprise.
BEAT $\rightarrow$ BUY (0.65 small, 0.85 if surprise $>$ \$0.10);
MISS $\rightarrow$ SELL (symmetric); in-line + optimistic MD\&A
$\rightarrow$ BUY 0.60; in-line + cautious $\rightarrow$ SELL 0.55;
momentum shift $\pm 0.10$. Confidence never below 0.40. Output JSON. &
10-Q (\{date\}, \{quarter\}): MD\&A sentiment, tone, forward-looking
language, business momentum, risk-factor delta, summary. \newline
IBES (\{date\}): consensus, actual, surprise, analyst count. \\
\midrule

\textbf{Strategy} &
You are a Strategy Specialist analyzing the most recent 10-K. Judge
long-term competitive trajectory. Optimistic + $\geq$3 growth signals
$\rightarrow$ BUY 0.65; optimistic + $\geq$2 growth + stable risks
$\rightarrow$ BUY 0.55; cautious + worsening risks $\rightarrow$ SELL
0.60; mixed but improving $\rightarrow$ BUY 0.45; mixed but stable
$\rightarrow$ HOLD 0.40. Output JSON. &
10-K filed: \{filed\_date\}, period: \{period\_end\} \newline
Business: sentiment, tone, top-5 themes, top-5 growth signals \newline
Risk: sentiment, tone, top-5 categories \newline
Overall summary (200 chars) \\
\midrule

\textbf{Fundamentals} &
You are a Fundamentals Specialist analyzing Compustat ratios. IGNORE
the P/E ratio (TSLA's is always extreme). Revenue growth $> +10\%$
QoQ $\rightarrow$ BUY 0.80; 0--10\% $\rightarrow$ BUY 0.55; declining
$\rightarrow$ SELL 0.65--0.85; operating-margin shift $\pm 0.10$;
negative-to-positive margins $\rightarrow$ STRONG BUY 0.85; ROE
$> 10\%$ + growth $\rightarrow$ BUY 0.75. Output JSON. &
\{quarter\} (reported \{rdq\}): Revenue + QoQ growth, EPS + QoQ
growth, margins (operating, net, gross), ROE, ROA, debt-to-equity,
current ratio. \\
\midrule

\textbf{Analyst} &
You are an Analyst Specialist analyzing IBES consensus estimates.
Recent BEAT $\rightarrow$ BUY (0.85 if surprise $>$ \$0.10, else
0.65); recent MISS $\rightarrow$ SELL (symmetric); analyst count
$> 20$ adds $+0.05$; tight estimate range ($<$ \$0.30) $\rightarrow$
BUY 0.65; wide range ($>$ \$0.50) $\rightarrow$ SELL 0.50. HOLD only
if literally no data. Output JSON. &
IBES (\{date\}): num analysts, mean estimate, median, stdev,
low--high range, last actual EPS, last surprise. \\
\midrule

\textbf{Technical} &
You are a Technical Specialist. MACD $>$ signal $\wedge$ price
$>$ SMA50 $\rightarrow$ BUY 0.75 (symmetric SELL); MACD bullish cross
alone $\rightarrow$ BUY 0.55; RSI $> 70$ $\rightarrow$ SELL 0.65;
RSI $< 30$ $\rightarrow$ BUY 0.65; RSI 40--60 + uptrend $\rightarrow$
BUY 0.60; ADX $> 25$ amplifies confidence $\pm 0.10$. Output JSON. &
TA (\{date\}): close, RSI(14) + overbought/oversold label, MACD +
signal + bullish/bearish-cross label, SMA20, SMA50 + up/downtrend
label, Bollinger lower/upper, ADX(14) + strong/weak label. \\
\midrule

\textbf{Social} &
You are a Social Sentiment Specialist analyzing r/wallstreetbets.
WSB = retail FOMO/panic; treat as signal, not contrarian. High
volume ($> 40$/wk) + bullish tone $\rightarrow$ BUY 0.55; high volume
+ panic $\rightarrow$ SELL 0.55; low volume ($< 20$) $\rightarrow$
HOLD 0.30; mixed + high volume $\rightarrow$ BUY 0.45; ``puts/short''
mentions $\rightarrow$ SELL 0.55; ``calls/moon/yolo'' $\rightarrow$
BUY 0.55. Output JSON. &
Total TSLA posts in last 7 days, avg score, avg comments, top-15
posts by score with titles (120 chars each). \\
\midrule

\textbf{Meta agent} &
You are the Chief Trading Officer. Eight specialists analyzed TSLA;
make a decisive final call. Weighting: NEWS (highest, may override
if conf $> 0.70$) $\succ$ Event $\succ$ Earnings $\succ$ Technical
$\approx$ Fundamentals $\approx$ Analyst $\succ$ Social $\succ$
Strategy. If $\geq$5 specialists agree, follow with high confidence.
HOLD only if truly split with no clear lean. Output JSON. &
Specialist verdicts (News first): \newline
\texttt{[NEWS] ACTION (conf=X.XX) -- reasoning} \newline
\texttt{[EVENT] ACTION (conf=X.XX) -- reasoning} \newline
\ldots \newline
Make the final TSLA trading decision. \\

\end{longtable}
\normalsize

\end{document}